\pdfoutput=1

\documentclass[11pt]{article}

\usepackage{acl}

\usepackage{times}
\usepackage{latexsym}

\usepackage[T1]{fontenc}

\usepackage[utf8]{inputenc}

\usepackage{microtype}
\usepackage{booktabs}
\usepackage{comment}
\usepackage{amsmath}
\usepackage{graphicx}
\usepackage{multirow}
\makeatletter
\def\thickhline{%
  \noalign{\ifnum0=`}\fi\hrule \@height \thickarrayrulewidth \futurelet
   \reserved@a\@xthickhline}
\def\@xthickhline{\ifx\reserved@a\thickhline
               \vskip\doublerulesep
               \vskip-\thickarrayrulewidth
             \fi
      \ifnum0=`{\fi}}
\makeatother

\newlength{\thickarrayrulewidth}
\title{\SM{}: A Large-Scale Mined Corpus \\
of Multilingual Speech-to-Speech Translations}

\author{
  Paul-Ambroise Duquenne$^{1,2,}$\thanks{~~Equal contributions} \and
  Hongyu Gong$^{1,*}$ \\
  \textbf{Ning Dong}$^1$ \and
  \textbf{Jingfei Du}$^1$ \\
  \textbf{Ann Lee}$^1$ \and
  \textbf{Vedanuj Goswani}$^1$ \and
  \textbf{Changhan Wang}$^1$ \\
  \textbf{Juan Pino}$^1$ \and \textbf{Beno\^it Sagot}$^2$ \and 
  \textbf{Holger Schwenk}$^1$ \\
  $^1$ Meta AI Research, $^2$ Inria \\
  \texttt{padqn,hygong,dnn,jingfeidu,schwenk@meta.com} 
}

\begin{document}

\newcommand{\todo}[1]{{\color{red}[\textbf{TODO:}#1]}}
\newcommand{\holger}[1]{{\color{blue}[\textbf{Holger:}#1]}}
\newcommand{\paul}[1]{{\color{green}[\textbf{Paul:}#1]}}
\newcommand{\hongyu}[1]{{\color{orange}[\textbf{Hongyu:}#1]}}
\newcommand{\juan}[1]{\textcolor{purple}{[JP: #1]}}
\newcommand{\ning}[1]{\color{pink}{[Ning: #1]}}
\newcommand{\cw}[1]{\textcolor{olive}{[CW: #1]}}
\newcommand{\SM}{\mbox{SpeechMatrix}}
\newcommand{\voxpop}{\mbox{VoxPopuli}}
\newcommand{\covost}{\mbox{CoVoST~2}}
\newcommand{\flores}{\mbox{FLoRes-101}}
\newcommand{\MC}{\multicolumn} 
\newcommand{\nbLangs}{17}    
\newcommand{\nbMinedPairs}{136}
\newcommand{\nbHavg}{1,537}    
\newcommand{\nbHtotal}{418k}   
\newcommand{\nbHtotalText}{418 thousand}
\newcommand{\nbSentsEN}{21.5 billion}   
\pagestyle{plain}
\newcommand{\tless}{\mbox{Textless model}}
\newcommand{\xm}{\mbox{XM Transformer}}

\maketitle
\begin{abstract}

We present \SM{}, a large-scale multilingual corpus of speech-to-speech translations mined from real speech of European Parliament recordings. It contains speech alignments in \nbMinedPairs{} language pairs with a total of \nbHtotalText{} hours of speech. To evaluate the quality of this parallel speech, we train bilingual speech-to-speech translation models on mined data only and establish extensive baseline results on Europarl-ST, \voxpop{} and FLEURS test sets. Enabled by the multilinguality of \SM{}, we also explore multilingual speech-to-speech translation, a topic which was addressed by few other works. We also demonstrate that model pre-training and sparse scaling using Mixture-of-Experts bring large gains to translation performance. The mined data and models are freely available.

\end{abstract}

\section{Introduction}

Research has progressed in the area of speech-to-speech translation (S2ST) with the goal of seamless communication among people who speak different languages. Direct S2ST models attract increasing research interest, e.g. \cite{DBLP:conf/interspeech/JiaWBMJCW19}. Compared to conventional cascaded models, direct models do not rely on intermediate text representations which make them applicable to the translation of languages without a well-defined writing script. 
Moreover, direct S2ST have the advantage of higher training and inference efficiency \cite{DBLP:conf/acl/LeeCWGPMPAHTPH22}.

Despite the benefits of direct approaches, model training is faced with the major issue of data scarcity. Human labeled speech data is expensive to create, there are very few data resources providing parallel speech, and the data amount is quite limited. 
To mitigate the data scarcity, some works have leveraged multitask learning \cite{DBLP:conf/interspeech/JiaWBMJCW19,DBLP:conf/acl/LeeCWGPMPAHTPH22},  data augmentation with speech variation \cite{DBLP:conf/interspeech/JiaWBMJCW19}, or with synthesized speech \cite{DBLP:journals/corr/abs-2203-13339,DBLP:journals/corr/abs-2204-02967}, and knowledge from pre-trained models \cite{DBLP:conf/naacl/LeeGDSCWPAPGH22,DBLP:journals/corr/abs-2204-02967} such as HuBERT \cite{DBLP:journals/taslp/HsuBTLSM21}, wav2vec 2.0 \cite{DBLP:conf/nips/BaevskiZMA20} and mBART \cite{liu2020multilingual}.

Recently, the multilingual speech/text sentence embedding space from \citet{duquenne2021multimodal} enabled the first speech mining results, aligning speech and text in different languages. Using this mined data to train direct speech-to-text and speech-to-speech translation systems can improve the performance of such models \cite{duquenne2021multimodal,DBLP:conf/naacl/LeeGDSCWPAPGH22}. 
Finally, \citet{tmodules} showed that such multilingual and multimodal sentence embeddings could be decoded into different languages and/or modalities in a zero-shot way, which suggests that multilingual speech content is well encoded in these fixed-size representations.

In this work, we trained speech encoders for \nbLangs{} languages\footnote{Czech (cs), German (de), English (en), Spanish (es), Estonian (et), Finnish (fi), French (fr), Croatian (hr), Hungarian (hu), Italian (it), Lithuanian (lt), Dutch (nl), Polish (pl), Portuguese (pt), Romanian (ro), Slovak (sk) and Slovenian (sl).} and mined speech-to-speech alignments for all possible language pairs. To the best of our knowledge, \SM{} is by far the largest freely available speech-to-speech translation corpus, with \nbMinedPairs{} language directions and an average of \nbHavg{} hours of source speech in each direction for a total of \nbHtotalText{} hours.
We demonstrate that strong S2ST models can be trained with these mined data and validate the good quality of the speech alignments across languages. We are open-sourcing the mined data, the speech encoders used for mining, multilingual HuBERT models in four language families for target unit generation, language-specific vocoders for speech synthesis from discrete units, and S2S models trained and presented in this work.\footnote{ \url{https://github.com/facebookresearch/fairseq/tree/ust/examples/speech_matrix}}

\section{Related Works}
\textbf{From bitext mining to speech mining}. State-of-the-art machine translation models are trained on labeled translation data, commonly called bitexts. In addition to human-labeled data, automatic methods have been proposed to find parallel sentences in a source and target language in monolingual resources. This task is known as bitext mining. In order to decide if sentences should be aligned, some works used document meta-information \citep{resnik1999mining}, cross-lingual document retrieval \citep{munteanu2005improving} or machine translation and information retrieval \citep{rauf2009comparable,bouamor2018h2}. More recent work use multilingual sentence embeddings to perform bitext mining, calculating cosine similarity \citep{Schwenk:2018:acl_mine} or other margin-based similarity \citep{Artetxe:2018:mine_arxiv,Yang:2019:ijcai_add_marg_softmax} in the embedding space to decide which sentences should be aligned. A large variety of methods has been explored to learn multilingual sentence embedding spaces \citep{espana2017empirical,Schwenk:2017:repl4nlp,Artetxe:2018:tacl_laser,Yang:2019:ijcai_add_marg_softmax,Reimers:2019:arxiv_sent_bert,Yang:2019:multilin_univ_sent_encod,Feng:2020:arxiv_labse}. Such methods enabled the creation of massively multilingual mined corpora of bitexts like the CCMatrix project \cite{schwenk:2021:acl_ccmatrix}. These bitexts were successfully used to train state-of-the-art machine translation models, e.g. \cite{fan:2020:jalt_m2m,deepnet:2022:arxiv}. Finally, existing sentence embedding spaces can be extended to new languages \citep{Reimers:2020:emnlp_xdestl,laser3} or the speech modality \cite{duquenne2021multimodal,khurana2022samu} with knowledge distillation, also called teacher-student approach. These multilingual and multimodal sentence embeddings enabled to perform large-scale speech-text mining, or speech-speech mining for a small set of languages.

\textbf{Speech-to-speech translation}. Early works on speech-to-speech translation are cascaded systems typically consisting of automatic speech recognition (ASR), machine translation (MT) and text-to-speech synthesis (TTS) \cite{DBLP:journals/taslp/NakamuraMNKKJZYSY06,DBLP:conf/iwslt/DoSNTN15}. The reliance on intermediate text outputs makes cascaded models unable to support unwritten languages. Moreover multiple separate components make the training and inference inefficient. Given these limitations, there has been a recent surge of research interest in direct approaches to speech translation without the need of texts. Translatotron \citep{DBLP:conf/interspeech/JiaWBMJCW19} is the first end-to-end S2ST model built upon a sequence-to-sequence architecture, which is trained to generate target spectrograms from source speech using multitask learning. As an improved version, Translatotron2 \cite{DBLP:conf/icml/JiaRRP22} has the ability to preserve voice in translated speech. It adopts two decoders: a linguistic decoder to predict phoneme and an acoustic decoder to predict target spectrograms. 

Another line of research replaces the target spectrograms in S2ST modeling with discrete units which are learned from a large amount of unlabeled speech \cite{DBLP:conf/acl/LeeCWGPMPAHTPH22,DBLP:conf/naacl/LeeGDSCWPAPGH22}. Discrete units are shown to better capture linguistic content than spectrograms, making the translations more robust to speech variations from different speakers or prosody \cite{DBLP:journals/taslp/HsuBTLSM21}.

Despite these progress on direct S2ST, direct approaches are faced with the challenge of limited parallel speech. We will describe existing speech-to-speech datasets in the next part. Data augmentation is a straightforward way of increasing the data such as speech transformation \cite{DBLP:conf/interspeech/JiaWBMJCW19} and speech synthesis via TTS \cite{DBLP:journals/corr/abs-2203-13339}.
Multitask learning is a commonly used technique to better train the model on small amount of labeled data \cite{DBLP:conf/icml/JiaRRP22,DBLP:conf/acl/LeeCWGPMPAHTPH22}. Another effective method is to leverage pre-trained components for model initialization.

\begin{table*}[t!]
\resizebox{1.0\textwidth}{!}{
\begin{tabular}{ccccc}
\toprule
Dataset & \# of Languages & Avg. duration (h) & Source speech & Target speech \\ \midrule
Fisher \cite{post2014fisher} & 2 & 127 & Telephone conversation & Synthetic \\
MaSS \cite{DBLP:conf/lrec/BoitoHGFB20} & 8 & 20 & Bible reading & Bible reading \\
\voxpop{} \cite{DBLP:conf/acl/WangRLWTHWPD20} & 15 & 82 & European Parliament speech & Simultaneous interpretation \\
CVSS (C+T) \cite{DBLP:journals/corr/abs-2201-03713} & 21 & 181 & Read & Synthetic \\
FLEURS \cite{DBLP:journals/corr/abs-2205-12446} & 102 & 12 & Read & Read \\ 
\SM{} (ours) & \nbLangs{} & 1537 & European Parliament speech & European Parliament speech \\ \bottomrule
\end{tabular}}
\caption{A comparison of existing speech-to-speech datasets.}
\label{tab:speech_corpora}
\end{table*}

\textbf{Speech translation corpora}.
The Fisher dataset, a collection of approximately $170$ hours of telephone conversations in Spanish \cite{post2014fisher}, is commonly used as training data for Spanish-English S2ST,  However it does not have parallel English speech. Previous works generate synthesized English speech from English text translations provided by Fisher. Another S2S dataset containing synthesized speech is CVSS which covers parallel S2ST translations from $21$ languages into English. It is derived from Common Voice \cite{DBLP:conf/lrec/ArdilaBDKMHMSTW20} and \covost{} \cite{wang2021covost}, and synthesizes speech from translated texts via TTS models.
The release of \voxpop{} dataset provided the largest S2S translations in real speech so far \cite{DBLP:conf/acl/WangRLWTHWPD20}. It covers pairwise speech-to-speech translations among $15$ languages, and each direction has less than $500$ hours of speech. 
In another initiative named FLEURS, the text-to-text evaluation data of the \flores{} benchmark \cite{DBLP:journals/tacl/GoyalGCCWJKRGF22} was extended to the speech modality.
Supporting $102$ languages, FLEURS has a larger language coverage than \voxpop{}, but it contains speech of only around $12$ hours per language and it is intended to be used as $N$-way parallel test set.


In this work, we present \SM{}, a large-scale multilingual speech-to-speech corpus mined from \voxpop{} \cite{DBLP:conf/acl/WangRLWTHWPD20}. It contains speech alignments in $\nbMinedPairs{}$ language pairs with an average of $1,537$-hour source speech per direction.
The main characteristics of these speech corpora are summarized in \autoref{tab:speech_corpora}.

\section{Speech-to-Speech Mining}

The mining approach used in this work is built upon the core idea of encoding multilingual speech utterances into a shared embedding space. Speech encoders project utterances with similar semantic content to fixed-size representations, the resulting embedding vectors are close in the embedding space regardless of their languages. The closeness of speech embeddings thus reflects the similarity of speech content, which serves as the speech alignment score in the mining process. In this section, we discuss speech encoders and speech alignment mining.

\subsection{Speech Encoders}
\begin{figure}[b!]
    \centering
    \includegraphics[width=0.45\textwidth]{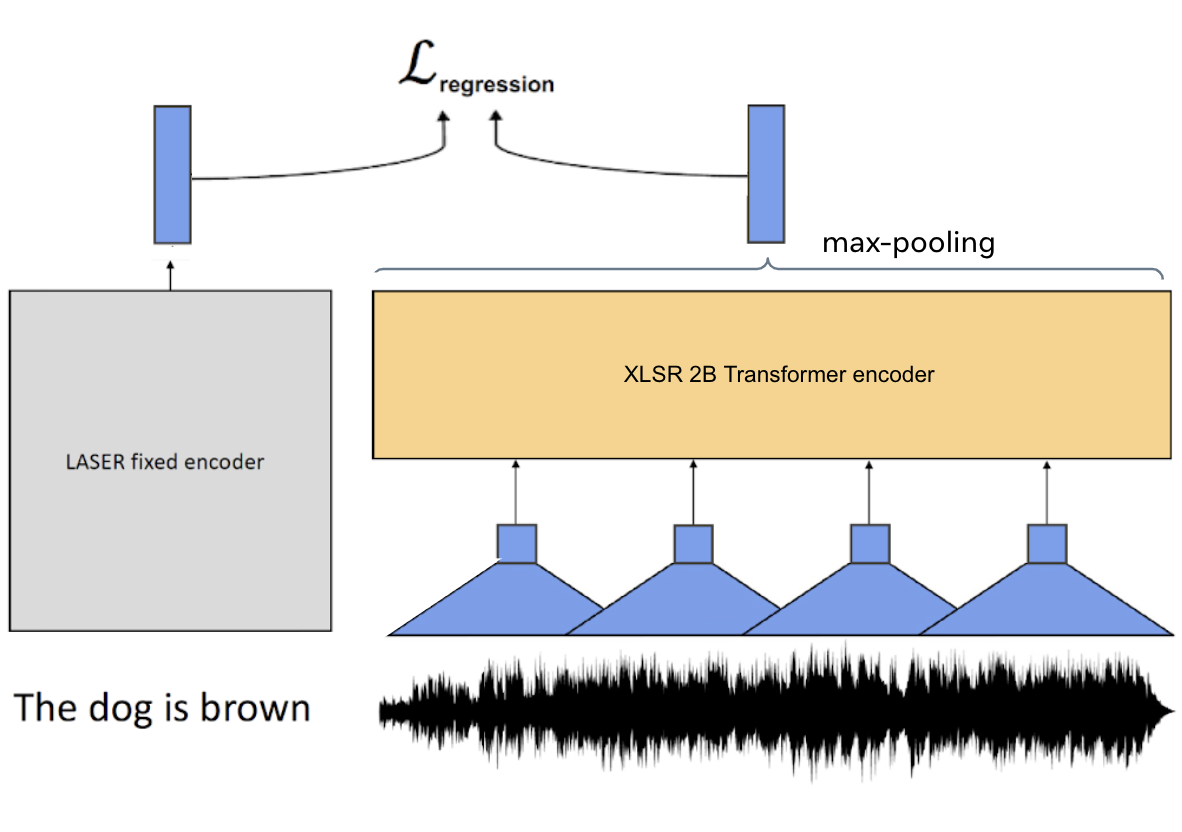}
    \caption{Architecture of speech encoders training.} 
    
    \label{fig:arch}
\end{figure}

We followed the teacher-student approach introduced in \cite{duquenne2021multimodal} and trained speech encoders with the supervision of the multilingual LASER text encoder \cite{DBLP:conf/acl/SchwenkWEGJF20}. The LASER text space has proven to have interesting semantic properties for mining purposes \cite{schwenk:2021:acl_ccmatrix,nllb}. In order to have similar semantic properties in the multilingual speech embedding space, the LASER text encoder is used as the teacher to train speech encoders. Transcriptions or written translation of the audio utterances are encoded with LASER text encoder as target vectors for speech encoders. During training, we minimize the cosine loss between fixed-size representations output by speech encoders, and the outputs of LASER text encoder (which weights are fixed during training). Speech encoders are initialized with the 2B parameter XLS-R model \cite{DBLP:journals/corr/abs-2111-09296}, which was pre-trained on nearly half a million hours of publicly available speech audio in 128 languages. Following \cite{tmodules}, the fixed-size representation for speech are obtained with max-pooling of the encoder outputs which appeared to work better compared to other pooling methods. We summarize the architecture of the speech encoder training in \autoref{fig:arch}.

\begin{table*}[t!]
\centering
\resizebox{1.0\textwidth}{!}{
\begin{tabular}{c|ccccccccccccccccc}
  \toprule
  \textbf{Sim Search} & \textbf{cs} & \textbf{de} & \textbf{en} & \textbf{es} & \textbf{et} & \textbf{fi} & \textbf{fr} & \textbf{hr} & \textbf{hu} & \textbf{it} &   \textbf{lt} & \textbf{nl} & \textbf{pl} & \textbf{pt} & \textbf{ro} & \textbf{sk} & \textbf{sl}  \\
  \midrule
  \# test sentences  & 1k  & 1.7k & 1.5k  & 1.4k & 47  & 0.4k  & 1.5k  & 0.3k  & 1k  & 1k  &  39 & 1k  & 1.6k & --- & 1.3k  & 0.6k  & 0.3k  \\
  \midrule
  Audio vs. transcriptions  &  0.6 &  1.0 & 0.2 &  0.7 & 0.0 & 0.7 &  0.5   & 0.3 &  1.1 & 4.9 &  0.0 & 0.8   &  0.9 &  --- & 0.9 & 0.7 & 3.1    \\
  \bottomrule
\end{tabular}}
\caption{Similarity search error rates (in \%) on \voxpop{} ASR test set.}
\label{simsearch_vp}
\end{table*}

\citet{duquenne2021multimodal} presented an ablation study on different text teacher choices: using transcription embeddings as target, written translation embeddings as target, or both. Their conclusion was that using both transcriptions and written translations embeddings as target yield the best similarity search results.
We used various publicly available ASR data sets which cover our languages to train the speech encoders, like \covost{} \cite{wang2020covost,wang2021covost}, Common Voice (citation), Europarl \cite{DBLP:conf/lrec/ArdilaBDKMHMSTW20}, mTedx \cite{DBLP:conf/interspeech/SaleskyWBCNTOP21}, Must-C \cite{di-gangi-etal-2019-must} or \voxpop{} \cite{DBLP:conf/acl/WangRLWTHWPD20}, as well as speech translation data from the foreign languages into English and from English into German.
We removed training samples where the transcription or the written translation consisted of multiple sentences, as LASER has been trained on single sentences only. For better training efficiency, we trained speech encoders for each language family instead of each language. The language grouping is provided in \autoref{tab:fam}. To better handle imbalanced training data, we sample the training data from different languages with the same method and hyper-parameter as \cite{duquenne2021multimodal}. For English (en), Slovenian (sl), Lithuanian (lt) and Dutch (nl), we also trained separate monolingual speech encoders that appeared to have lower valid cosine loss compared to multilingual speech encoders, and these ones were used for mining.

\subsection{Evaluation of speech encoders}

Similarity search is frequently used to evaluate multilingual text encoders, e.g. \citep{Artetxe:2018:mine_arxiv,Feng:2020:arxiv_labse,laser3}. 
The underlying approach is to first calculate the embeddings of all source and reference sentences in a test set. Then, for each source sentence, the closest embedding among all reference sentences is searched, if it is not the expected one, an error is counted. 
Following \citet{duquenne2021multimodal}, we extend similarity search to the multimodal setting and use the following score to measure similarity between the source audio, and the target transcriptions or translations:
\begin{eqnarray}
  \label{eqn:margin}
  sim(x,y) \hspace*{-1.5cm} && \\
  \nonumber
    \hspace*{-2cm} & = & cos(x, y) \\
    \nonumber
    \hspace*{-2cm} && - \left(
    \sum\limits_{z \in NN_k(x)}\hspace{-10pt}\cfrac{cos(x, z)}{2k}
    + \sum\limits_{z \in NN_k(y)}\hspace{-10pt}\cfrac{cos(y, z)}{2k}
    \right)
\end{eqnarray}

\noindent where $x$ and $y$ are the source and target embeddings, and  $NN_k(x)$ denotes the $k$ nearest neighbors of $x$. We used $k=4$.
%
%
%
We evaluated similarity search of audio against transcriptions on \voxpop{} ASR test set in \autoref{simsearch_vp}, which is our target domain as we plan to mine unlabeled speech from \voxpop{} (see  \autoref{large_scale_mining}). We also evaluated similarity search of audio against written translations or transcriptions on \covost{} test set in \autoref{simsearch_cv}, in order to compare with previous work for German, English, Spanish and French (details can be found in the Appendix). Finally, we report text-to-text similarity search using the LASER text encoder as lower bound for the speech translation similarity search error rate since we use gold transcriptions to search against written translations. We report error rates (in \%) that correspond to the percentage of audio utterances wrongly matched with text transcripts from the same test set. We notice that error rates are very low for all languages (below 5\% and around 1 or 2\% for most languages), which is an initial validation of the quality if the speech encoders before large-scale mining.


\begin{table}
\centering
\resizebox{0.48\textwidth}{!}{
\begin{tabular}{l|ccccccccc}
\thickhline
 & \textbf{de} & \textbf{en} & \textbf{es} & \textbf{et}  & \textbf{fr} & \textbf{it}  & \textbf{nl} & \textbf{pt} &  \textbf{sl}  \\
\hline
\# test sentences &  14k & 16k  & 13k  & 2k & 15k  & 9k & 2k & 4k & 0.4k \\
\hline
\multicolumn{1}{l|}{\textbf{Audio}} & \multicolumn{9}{l}{}\\
~~~ vs. transcriptions  & 1.4    & 2.9    & 0.4    & 0.1          & 0.5      & 0.5          & 1.0       & 1.1      & 1.7    \\
~~~ vs. en translations         & 3.3         &   ---          & 1.3        & 1.0               & 1.5        &      1.7               & 4.4               &        1.9   &        4.4         \\
\multicolumn{1}{l|}{\textbf{Text transcription}} & \multicolumn{9}{l}{}\\
~~~ vs. en translations        & 2.0         &    ---         & 1.0         & 0.1              & 1.0        &     1.3               & 2.4             &   0.7          &       0.8  \\       
\thickhline
\end{tabular}
}
\caption{Similarity search error rates (in \%) on \covost{} test set.}
\label{simsearch_cv}

\end{table}

\subsection{Large-scale speech mining}
\label{large_scale_mining}
We used \voxpop{} as our source of unlabeled speech for the \nbLangs{} languages of focus. We downloaded the full unsegmented parliament session recordings from \voxpop{} github repository.\footnote{\url{https://github.com/facebookresearch/voxpopuli}} We present in \autoref{unlabeled} the number of hours of unlabeled speech for each language.

In principle, performing speech-to-speech or speech-to-text mining can be done with exactly the same pipeline than text-to-text mining.
The calculation of the embeddings uses of course different encoders, but the rest of the pipeline is exactly the same.
We follow the global mining approach as described in \citet{schwenk:2021:acl_ccmatrix} and compare all segments in the source language with all segments in the target language. Similarity scores are calculated in both directions using the margin as described in \autoref{eqn:margin} considering $k=16$ neighbors. Segments are considered to be parallel if the margin score exceeds a threshold, we use 1.06 if not specified otherwise. The reader is referred to \citet{schwenk:2021:acl_ccmatrix} for a detailed description of the generic mining pipeline.

\begin{table}[t!]
\centering\small
\begin{tabular}{c|r||c|r||c|r}
\thickhline
Lang & Hours & Lang & Hours & Lang & Hours \\
\hline
 cs   & 18.7k & fr   & 22.8k & pl   & 21.2k \\
 de   & 23.2k & hr   & 8.1k  & pt   & 17.5k \\
 en   & 24.1k & hu   & 17.7k & ro   & 17.9k \\
 es   & 21.4k & it   & 21.9k & sk   & 12.1k \\
 et   & 10.6k & lt   & 14.4k & sl   & 11.3k \\
 fi   & 14.2k & nl   & 19.0k & \\
\thickhline
\end{tabular}
\caption{Number of hours of raw unlabeled speech from \voxpop{} by language.}
\label{unlabeled}
\end{table}


There is however one important difference when processing speech: it is not straight-forward to segment the audio signal into parts which have the optimal granularity for mining. When aligning texts, the monolingual data is usually segmented into sentences, since a sentence is a well defined semantic concept and there are reliable algorithms to perform this sentence segmentation.
The \voxpop{} recordings have a rather long duration, e.g. one hour and a half in average for English. We apply Voice Activity Detection (VAD) using Silero-VAD\footnote{\url{https://github.com/snakers4/silero-vad}}  which supports over 100 languages. The resulting segments do not necessarily correspond to sentences. On one hand, there may be a silence in the middle of a sentence, e.g. a hesitation. And on the other hand, two sentences may follow each other without a long silence separating them.
We follow the ``over segmentation approach'' outlined in \citet{duquenne2021multimodal}: several possible segments are created and we let the mining algorithm decide which ones match best. Initial experiments suggest that sentences shorter than 1 sec or longer than 20 sec are unlikely to be aligned and were excluded. 

After mining, the resulting speech alignments may have overlap as we over-segmented the unlabeled speech. \citet{duquenne2021multimodal} introduced a post-processing method to remove overlaps between mined speech segments on the source speech side. We relax a little bit the post-processing of the mined data, allowing for some overlap between mined speech segments: for two audio segments that overlap on the source side, if the overlap represent more than 20\% of the first segment and of the second segment, we discard the alignment with the lowest mining score. We did an ablation study on different percentage thresholds for one low resource, one mid-resource and one high-resource pair and found that 20\% was the best threshold for all settings. 

We report the statistics of the mined speech-to-speech translation pairs in \autoref{stats}, with a mining score threshold of $1.06$. The mined data totals \nbHtotal{} hours of parallel speech with an average of \nbHavg{} hours of source speech on all translation directions. While some high resource languages like English (en), Spanish (es) or French (fr) can reach up to 5k hours of aligned speech with other spoken languages; lower resource languages such as Estonian (et), Croatian (hr), Slovenian (sl) and Lithuanian (lt) obtain much less aligned speech, with only a few hours of aligned speech for Lithuanian.

We also performed mining of the source speech in sixteen languages against more than twenty billion English sentences from Common Crawl. This yielded between 827 and 3966 hours of speech-text alignments (see the last column of \autoref{stats}). Training and evaluation of speech-to-text translation is left for future research. An important advantage of our teacher-student approach is that all speech and text encoders are derived from the same LASER teacher and are mutually compatible. This enables us to perform speech-to-text mining for many more languages \citep{nllb}.

\begin{table*}[htbp!]
\centering
\resizebox{1.0\textwidth}{!}{
\begin{tabular}{c|*{17}{r}|r}
\thickhline
& \MC{17}{c|}{\textbf{Speech targets}} & \textbf{Text} \\
\textbf{Src/Tgt} & \textbf{cs} & \textbf{de} & \textbf{en} & \textbf{es} & \textbf{et} & \textbf{fi} & \textbf{fr} & \textbf{hr} & \textbf{hu} & \textbf{it} & \textbf{lt} & \textbf{nl} & \textbf{pl} & \textbf{pt} & \textbf{ro} & \textbf{sk} & \textbf{sl} & \textbf{en}\\ \hline
\textbf{cs} & - & 2381 & 3208 & 2290 & 952 & 1312 & 2476 & 726 & 1396 & 2410 & 84 & 2377 & 2516 & 1867 & 1190 & 2146 & 452  & 2528\\ 
\textbf{de} & 2386 & - & 4734 & 3113 & 901 & 1477 & 3536 & 498 & 1871 & 3476 & 41 & 3384 & 2632 & 2250 & 1281 & 1646 & 361 & 3073 \\ 
\textbf{en} & 3172 & 4676 & - & 4715 & 1585 & 2169 & 5178 & 824 & 2266 & 4897 & 82 & 4422 & 3583 & 3572 & 2258 & 2306 & 586 & - \\ 
\textbf{es} & 2240 & 3041 & 4708 & - & 862 & 1373 & 4446 & 528 & 1599 & 4418 & 47 & 3067 & 2646 & 3484 & 1857 & 1603 & 308 & 3966\\ 
\textbf{et} & 943 & 892 & 1593 & 877 & - & 1201 & 934 & 265 & 1119 & 1019 & 39 & 1055 & 949 & 721 & 419 & 780 & 196  & 1578\\ 
\textbf{fi} & 1296 & 1463 & 2180 & 1393 & 1197 & - & 1449 & 306 & 1473 & 1599 & 47 & 1654 & 1350 & 1128 & 621 & 977 & 260 & 1969\\ 
\textbf{fr} & 2424 & 3457 & 5171 & 4455 & 923 & 1435 & - & 560 & 1711 & 4618 & 50 & 3273 & 2822 & 3384 & 1991 & 1657 & 326  & 3966\\ 
\textbf{hr} & 736 & 507 & 854 & 553 & 273 & 317 & 588 & - & 328 & 615 & 24 & 546 & 660 & 433 & 277 & 586 & 136 & 1311\\ 
\textbf{hu} & 1417 & 1897 & 2346 & 1672 & 1140 & 1507 & 1787 & 328 & - & 1855 & 68 & 1839 & 1566 & 1315 & 808 & 1064 & 311 & 2301\\ 
\textbf{it} & 2404 & 3460 & 4948 & 4500 & 1028 & 1614 & 4700 & 607 & 1823 & - & 103 & 3414 & 2848 & 3421 & 1995 & 1656 & 474 & 2891\\ 
\textbf{lt} & 78 & 38 & 79 & 46 & 37 & 44 & 48 & 21 & 61 & 95 & - & 77 & 80 & 35 & 18 & 64 & 6  & 827\\ 
\textbf{nl} & 2322 & 3305 & 4396 & 3066 & 1040 & 1633 & 3269 & 521 & 1768 & 3355 & 80 & - & 2459 & 2399 & 1352 & 1646 & 458 & 2708\\ 
\textbf{pl} & 2530 & 2646 & 3662 & 2735 & 967 & 1378 & 2913 & 656 & 1554 & 2883 & 88 & 2540 & - & 2121 & 1301 & 1892 & 431  & 2871\\ 
\textbf{pt} & 1849 & 2224 & 3606 & 3525 & 722 & 1131 & 3421 & 421 & 1279 & 3403 & 37 & 2436 & 2087 & - & 1579 & 1358 & 247 & 3540\\ 
\textbf{ro} & 1187 & 1275 & 2290 & 1894 & 423 & 627 & 2024 & 271 & 789 & 1996 & 19 & 1384 & 1288 & 1592 & - & 870 & 125 & 2784\\ 
\textbf{sk} & 2127 & 1628 & 2329 & 1631 & 781 & 982 & 1685 & 574 & 1038 & 1650 & 69 & 1676 & 1869 & 1361 & 867 & - & 370  & 2090\\ 
\textbf{sl} & 436 & 350 & 579 & 307 & 192 & 254 & 324 & 128 & 295 & 461 & 6 & 454 & 413 & 241 & 121 & 359 & -  & 1267\\ \thickhline
\end{tabular}}
\caption{Duration statistics (hours of source speech) of aligned speech-to-speech data between each pair of $\nbLangs{}$ languages (for mining threshold of $1.06$). The last column provides statistics for alignments of source speech against \nbSentsEN{} sentences of English texts.}
\label{stats}
\end{table*}

\subsection{Evaluation Data}

Besides the speech-to-speech data mined as the train set, we also leverage labeled public speech datasets as the evaluation sets.

\textbf{Test set}. In our experiments, we derive test sets in speech translation from three public corpora, evaluating translation models trained on mined data across different domains.

\begin{itemize}
    \item Europarl-ST (EPST) \cite{DBLP:conf/icassp/Iranzo-SanchezS20}. It is a multilingual speech-to-text translation corpus built from recordings of debates from the European Parliament, containing $72$ translation directions in $9$ languages.\footnote{en, fr, de, it, es, pt, pl, ro and nl}
    \item \voxpop{} \cite{DBLP:conf/acl/WangRLWTHWPD20}. S2S data, as part of \voxpop{} release, provides aligned source and target speech together with source transcription. We prepare the speech-to-text data with target speech and source transcription as our test set. To ensure that there is no overlap between the mined data and \voxpop{} test sets, we need to remove speech from mined alignments which are from the same session as test samples. In order to keep as much mined data as possible, we use \voxpop{} test set only when a language direction is not covered by EPST considering their domain similarity. Moreover, similarity scores are provided to indicate the quality of \voxpop{} samples. To choose high-quality data, we sort all sessions in the \voxpop{} S2S data in a decreasing order of the average similarity score of their samples. We keep adding samples from highly ranked sessions to the test set until the test size hits $1000$.
    \item FLEURS \cite{DBLP:journals/corr/abs-2205-12446}. Built upon N-way text translations from FLoRes \cite{DBLP:journals/tacl/GoyalGCCWJKRGF22}, FLEURS provides speech for these texts and creates speech-to-speech data covering all mined directions. We take its source speech and target texts as the test data. In the case where multiple utterances correspond to one piece of source text, we generate one test pair for each source utterance respectively. FLEURS texts are from English Wikipedia, which is a different domain from \voxpop{} and EPST.
\end{itemize}

\textbf{Valid set}. Valid sets are prepared using \voxpop{} and FLEURS data in a similar way as test sets. For \voxpop{}, we extract a valid set of about $1000$ samples by adding data from highly scored sessions which are not in the test set. FLEURS valid set is derived from its valid samples. We prepare speech-to-unit data from these selected valid samples by transforming the target speech into target units for speech-to-unit modeling which will be discussed in \autoref{sec:exp}.

\section{Experiments \& Results}
\label{sec:exp}


To evaluate the quality of the mined data, we trained S2ST models on \SM{} data and report the translation performance. We hope that these results will serve as baselines for future studies in speech translation.

\subsection{Experimental Setup}

\begin{figure}[htbp!]
    \centering
    \includegraphics[width=0.45\textwidth]{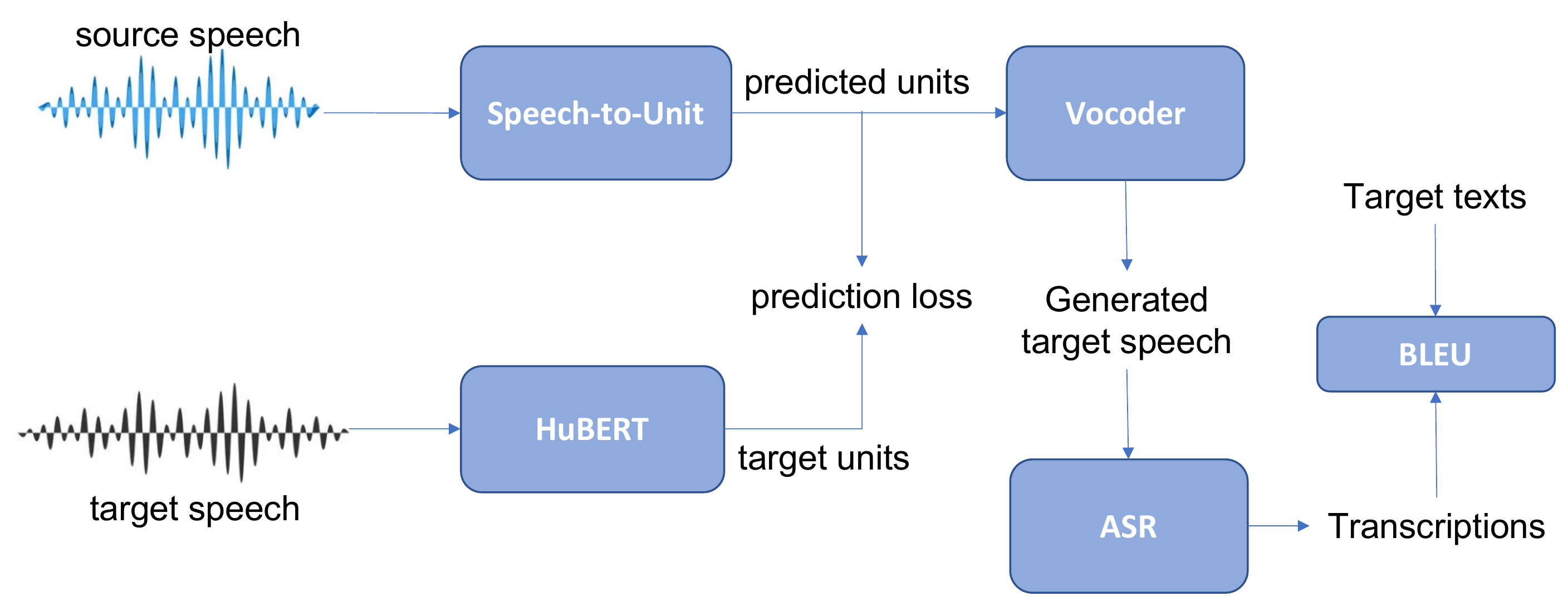}
    \caption{A Pipeline of Speech-to-Speech Translation and Evaluation.}
    \label{fig:pipeline}
\end{figure}

The training and evaluation pipeline of speech-to-speech translation is shown in \autoref{fig:pipeline}. Recent progress in speech-to-speech translation modeling suggests to discretize the target speech waveform into a unit sequence, relieving models from the complexity of predicting continuous waveform values. We borrow the idea of training speech-to-unit (S2U) model where units are pre-generated from target speech with a pre-trained HuBERT model \cite{DBLP:conf/acl/LeeCWGPMPAHTPH22}. During S2U training, models are periodically evaluated on the valid set of speech-to-unit samples, and the best checkpoint with the lowest valid loss is saved for model inference. 


When it comes to inference, speech could be synthesized from the predicted units with a vocoder, as the output of the S2S pipeline. It is then transcribed into texts by an off-the-shelf ASR model. The BLEU score is calculated by comparing the transcriptions against the ground truth target texts, which serves as the quantitative metric of mined data quality. We note that the ASR BLEU score is not a perfect metric for data quality, as it is unavoidably affected by the quality of ASR models. 
Next we discuss each module of the pipeline.

\textbf{Speech-to-Unit}. The S2U model takes the source speech and predicts a sequence of target units. It typically has an encoder-decoder architecture, where the encoder consists of convolutional and Transformer encoder layers, and the decoder is a Transformer decoder. We have experimented with different model variants, and discuss bilingual and multilingual training in  \autoref{sec:blg} and \autoref{sec:mlg} respectively.

\textbf{HuBERT}.  We reuse the same HuBERT model and $k$-means clusters for English, Spanish and French as in \cite{DBLP:conf/naacl/LeeGDSCWPAPGH22} for a fair comparison with existing results.  We also train HuBERT models for each language family as shown in \autoref{tab:fam} to cover other languages in \SM{}. We collect unlabeled \voxpop{} speech for all languages of the same family as the training data. Each HuBERT model is trained for three iterations, and more details of HuBERT training can be found in Appendix \ref{app:hubert}. We select the best layer for speech feature extraction and the best label size (i.e., the number of $k$-means clusters) using speech resynthesis together with the vocoder, which is discussed in Appendix \ref{app:vocoder}. With the optimal layer and label size decided, we could generate the target unit sequence for the given source speech.

\begin{table}[htbp!]
\centering\small
\begin{tabular}{cc}
\toprule
\bf Family & \bf Languages \\ \midrule
Romance & es, fr, it, pt, ro \\ 
Slavic & cs, pl, sk, sl, hr, lt \\ 
Germanic & en, de, nl \\ 
Uralic & fi, et, hu \\ \bottomrule
\end{tabular}
\caption{Language families in \voxpop{} data.}
\label{tab:fam}
\end{table}

\textbf{Vocoder}. Unit-based HiFi-GAN vocoders are trained to synthesize speech from unit sequence \cite{polyak21_interspeech}. In our experiments, vocoders are separately trained from S2U model.  We train vocoders on three datasets:
\begin{itemize}
    \item CSS10 \cite{DBLP:conf/interspeech/ParkM19}. CSS10 is a single-speaker corpus which we use to train vocoders in German, Finnish, Hungarian and Dutch. 
    \item \voxpop{} \cite{DBLP:conf/acl/WangRLWTHWPD20}. Given the ASR data with speaker information, we sort speakers based on their speech duration, and keep adding the top speakers until the speech is more than $20$ hours.
    \item Common Voice \cite{DBLP:conf/lrec/ArdilaBDKMHMSTW20}. Two languages, Portuguese and Estonian, are not covered by the two corpora above, and thus we turn to Common Voice. 
    Again, we select top speakers and prepare 12-hour and 10-hour speech for the vocoder training in Portuguese and Estonian respectively. 
\end{itemize}

We applied a denoiser\footnote{\url{https://github.com/facebookresearch/denoiser}} \cite{defossez2020real} to the speech of \voxpop{} and Common Voice as the speech preprocessing to increase signal-to-noise ratio (SNR) given that they are noisier than CSS10 audios. Then we prepare vocoder labels with HuBERT models generating $k$-means cluster labels for each utterance.  Single-speaker vocoders are trained in CSS10, and languages from \voxpop{} and Common Voice have multi-speaker vocoders where speaker embeddings are learned. During inference, we select the speaker with the longest speech duration to synthesize speech from predicted unit sequences, who has the most data for the vocoder to learn good speaker embeddings.  

\textbf{ASR}. We use off-the-shelf ASR models to transcribe the speech generated by vocoders, and details about the ASR models and their benchmark results are provided in the \autoref{app:asr}.

\section{Bilingual Speech-to-Speech Baselines}
\label{sec:blg}

In this part, we discuss the bilingual S2S models trained in each of $272$ language directions in \SM{}. The architecture of \tless{} is used for bilingual translation in our experiments \cite{DBLP:conf/acl/LeeCWGPMPAHTPH22}. A \tless{} consists of a speech encoder with $2$ convolution layers and $12$ Transformer encoder layers. Transformer layer has the embedding dimension of $512$ and the forward dimension of $2048$. It has two unit decoders with $6$ and $2$ Transformer decoder layers for target and source unit prediction respectively. The target unit decoder has the embedding dimension of $512$ and the forward dimension of $2048$, and the source unit decoder's dimensions are $256$ and $2048$.

\textbf{Training}. For a given direction, we extract units for source and target speech with their corresponding HuBERT models \cite{DBLP:journals/taslp/HsuBTLSM21}. 
Taking source speech, the model is trained to predict target unit sequence with cross-entropy loss as well as to reconstruct source units as an auxiliary task. 

\begin{figure}
    \centering
    \includegraphics[width=0.45\textwidth]{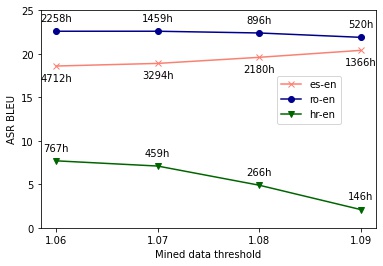}
    \caption{Bilingual S2S BLEU by mined data at different thresholds.} 
    \label{fig:threshold}
\end{figure}

For the sake of good translation performance and training efficiency, it is important to set a reasonable threshold to mined data and select samples above the threshold as the train set. We performed an analysis of translation performance varying with thresholds from $1.06$ to $1.09$ on three language pairs: es-en, ro-en and hr-en. \autoref{fig:threshold} shows the threshold, the corresponding speech data size and BLEU score. 

For low-resource directions such as hr-en, it is best to include all the mined data. For high-resource directions, es-en and ro-en, the optimal amount of mined data is around $1$k hours and it does not bring further gains to increase the data size. Given these observations, we choose the highest threshold that keeps the source speech duration in mined data more than $1$k hour for each direction. For example, we use a threshold of $1.09$ for es-en and of $1.06$ for hr-en.

\textbf{Comparison with existing results}. Since we adopt the same model as the previous work \cite{DBLP:conf/acl/LeeCWGPMPAHTPH22} and the difference only lies in the train set, it is straightforward to compare with its reported S2S translation results. 
\autoref{tab:comp} shows the results of S2ST models which are trained on our \SM{} mined data compared to \voxpop{} S2S data in each of four language directions: es-en, fr-en, en-es and en-fr.
The threshold of mined data is set as $1.09$ to these four directions, yielding an average of $1,436$-hour train set. Compared with $480$-hour labeled speech from \voxpop{}, \SM{} achieves an an average improvement of 5.4 BLEU, indicating the good quality and usefulness  of the mined data.

\begin{table}[t!]
\resizebox{0.48\textwidth}{!}{
\begin{tabular}{cccccc}
\toprule
Train set & & Es-En & Fr-En & En-Es & En-Fr \\
\midrule
\multirow{2}{*}{\begin{tabular}[c]{@{}c@{}}\voxpop{}\\ S2S\end{tabular}} & Hours & 532 & 523 & 415 & 451 \\
 & BLEU & 13.1  & 15.4  & 16.4  & 15.8  \\ \midrule
\multirow{2}{*}{\begin{tabular}[c]{@{}c@{}}\SM{}\\ ($t=1.09$)\end{tabular}} & Hours & 1,353 & 1,507 & 1,366 & 1,518 \\
& BLEU & 20.4  & 20.7  & 21.9  & 19.3  \\
\bottomrule
\end{tabular}}
\caption{BLEU scores on EPST test sets by S2ST models with different training data.}
\label{tab:comp}
\end{table}

\subsection{Large-Scale Bilingual Evaluation}

\begin{table*}[htbp!]
\centering
\resizebox{1.0\textwidth}{!}{
\begin{tabular}{c|ccccccccccccccccc}
\thickhline
 & \textbf{cs} & \textbf{de} & \textbf{en} & \textbf{es} & \textbf{et} & \textbf{fi} & \textbf{fr} & \textbf{hr} & \textbf{hu} & \textbf{it} & \textbf{lt} & \textbf{nl} & \textbf{pl} & \textbf{pt} & \textbf{ro} & \textbf{sk} & \textbf{sl} \\ \hline
\textbf{ cs } & - / - & 12.9 / 2.0 & 22.7 / 4.2 & 16.7 / 4.6 & - / 0.1 & 0.6 / 0.2 & 21.1 / 7.5 & 4.4 / 2.1 & 0.5 / 0.2 & 10.2 / 2.5 & 0.1 / 0.1 & 6.1 / 1.0 & 8.5 / 2.3 & - / 2.8 & 4.3 / 1.4 & 16.9 / 3.5 & 3.0 / 1.7 \\ 
\textbf{ de } & 7.3 / 2.3 & - / - & \underline{16.3} / 8.3 & \underline{11.7} / 3.8 & - / 0.1 & 1.2 / 0.2 & \underline{10.7} / 6.5 & 4.5 / 2.2 & 0.6 / 0.2 & \underline{3.8} / 1.8 & 0.1 / 0.0 & \underline{10.4} / 1.2 & \underline{3.5} / 0.9 & \underline{7.1} / 3.1 & \underline{5.2} / 2.1 & 3.0 / 0.8 & 4.1 / 1.0 \\ 
\textbf{ en } & 8.2 / 2.7 & \underline{10.1} / 2.7 & - / - & \underline{21.9} / 6.0 & - / 0.7 & 1.9 / 0.6 & \underline{19.2} / 10.4 & 8.4 / 2.4 & 1.1 / 0.3 & \underline{11.5} / 3.6 & 0.3 / 0.1 & \underline{15.1} / 3.8 & \underline{8.2} / 1.3 & \underline{11.8} / 5.1 & \underline{7.6} / 2.0 & 5.7 / 1.2 & 5.5 / 1.2 \\ 
\textbf{ es } & 5.2 / 1.9 & \underline{6.1} / 1.8 & \underline{20.4} / 7.5 & - / - & - / 0.1 & 1.3 / 0.2 & \underline{16.3} / 9.2 & 3.6 / 1.0 & 0.7 / 0.2 & \underline{11.1} / 4.2 & 0.1 / 0.1 & \underline{8.0} / 1.5 & \underline{3.9} / 1.4 & \underline{13.3} / 5.9 & \underline{5.2} / 2.3 & 2.2 / 0.9 & 2.2 / 0.8 \\ 
\textbf{ et } & - / 2.1 & - / 0.7 & - / 8.2 & - / 3.0 & - / - & - / 0.7 & - / 6.3 & - / 1.0 & - / 0.7 & - / 2.3 & - / 0.1 & - / 1.5 & - / 1.2 & - / 1.7 & - / 1.4 & - / 0.4 & - / 0.8 \\ 
\textbf{ fi } & 3.0 / 1.5 & 9.0 / 0.9 & 19.7 / 5.5 & 11.4 / 3.8 & - / 0.5 & - / - & 14.1 / 6.2 & 1.5 / 0.5 & 0.0 / 0.0 & 5.8 / 1.2 & 0.1 / 0.0 & 6.6 / 0.8 & 4.5 / 1.2 & - / 2.0 & 4.4 / 1.1 & 1.7 / 0.7 & 1.6 / 0.7 \\ 
\textbf{ fr } & 5.4 / 1.5 & \underline{6.3} / 2.1 & \underline{20.7} / 9.8 & \underline{18.4} / 7.6 & - / 0.1 & 0.8 / 0.2 & - / - & 5.4 / 1.7 & 0.7 / 0.2 & \underline{10.2} / 3.1 & 0.1 / 0.1 & \underline{8.4} / 1.3 & \underline{4.8} / 1.5 & \underline{13.4} / 5.8 & \underline{5.6} / 2.4 & 1.6 / 0.6 & 1.5 / 0.6 \\ 
\textbf{ hr } & - / 2.5 & - / 0.9 & - / 7.7 & - / 3.1 & - / 0.2 & - / 0.1 & - / 5.8 & - / - & - / 0.2 & - / 1.1 & - / 0.0 & - / 0.9 & - / 1.1 & - / 2.0 & - / 0.6 & - / 0.9 & - / 0.8 \\ 
\textbf{ hu } & 2.6 / 1.3 & 7.3 / 1.0 & 15.3 / 4.6 & 9.5 / 3.0 & - / 0.1 & 0.7 / 0.2 & 13.8 / 5.7 & 1.9 / 0.7 & - / - & 6.3 / 1.2 & 0.1 / 0.0 & 3.0 / 0.1 & 1.6 / 0.4 & - / 2.3 & 2.4 / 0.9 & 0.9 / 0.2 & 1.2 / 0.3 \\ 
\textbf{ it } & 6.4 / 1.3 & \underline{4.9} / 1.0 & \underline{18.9} / 6.3 & \underline{19.6} / 8.3 & - / 0.1 & 0.4 / 0.1 & \underline{15.3} / 11.3 & 5.2 / 1.3 & 0.7 / 0.2 & - / - & 0.1 / 0.0 & \underline{6.5} / 0.9 & \underline{3.6} / 1.1 & \underline{12.4} / 5.6 & \underline{3.7} / 1.9 & 2.1 / 0.4 & 2.8 / 0.6 \\ 
\textbf{ lt } & 0.2 / 0.1 & 0.0 / 0.0 & 3.1 / 0.9 & 0.8 / 0.2 & - / 0.0 & 0.0 / 0.0 & 0.7 / 0.2 & 0.1 / 0.0 & 0.0 / 0.0 & 0.6 / 0.4 & - / - & 0.7 / 0.1 & 0.1 / 0.0 & - / 0.0 & 0.0 / 0.0 & 0.0 / 0.0 & 0.1 / 0.0 \\ 
\textbf{ nl } & 3.5 / 1.4 & \underline{8.1} / 3.1 & \underline{18.0} / 5.7 & \underline{13.2} / 4.9 & - / 0.2 & 0.5 / 0.2 & \underline{13.0} / 7.5 & 3.3 / 1.8 & 0.4 / 0.2 & \underline{5.2} / 1.7 & 0.1 / 0.0 & - / - & \underline{3.4} / 0.9 & \underline{6.7} / 3.3 & \underline{4.1} / 1.4 & 1.7 / 0.4 & 2.1 / 1.0 \\ 
\textbf{ pl } & 7.2 / 1.6 & \underline{2.8} / 1.6 & \underline{4.9} / 4.9 & \underline{6.3} / 4.4 & - / 0.1 & 1.0 / 0.2 & \underline{5.5} / 5.4 & 4.5 / 1.2 & 0.5 / 0.1 & \underline{5.8} / 1.5 & 0.2 / 0.0 & \underline{1.6} / 0.3 & - / - & \underline{6.1} / 2.5 & \underline{3.2} / 1.2 & 4.7 / 1.1 & 2.4 / 0.7 \\ 
\textbf{ pt } & - / 1.2 & \underline{4.7} / 1.0 & \underline{21.2} / 6.1 & \underline{23.2} / 8.7 & - / 0.1 & - / 0.3 & \underline{18.1} / 11.1 & - / 1.1 & - / 0.1 & \underline{4.4} / 1.1 & - / 0.1 & \underline{5.0} / 0.6 & \underline{3.6} / 0.8 & - / - & \underline{4.4} / 1.5 & - / 0.6 & - / 0.6 \\ 
\textbf{ ro } & 4.6 / 1.9 & \underline{6.5} / 2.2 & \underline{22.6} / 7.8 & \underline{20.1} / 7.0 & - / 0.4 & 0.8 / 0.3 & \underline{18.6} / 11.3 & 2.4 / 0.9 & 0.4 / 0.2 & \underline{8.7} / 3.8 & 0.1 / 0.1 & \underline{3.5} / 0.9 & \underline{4.6} / 1.1 & \underline{10.3} / 6.0 & - / - & 2.3 / 0.7 & 0.7 / 0.2 \\ 
\textbf{ sk } & 28.2 / 9.1 & 10.7 / 2.1 & 21.4 / 5.5 & 15.5 / 5.1 & - / 0.3 & 1.0 / 0.2 & 19.2 / 7.8 & 5.0 / 3.0 & 0.5 / 0.4 & 4.7 / 2.1 & 0.1 / 0.0 & 4.2 / 0.7 & 5.3 / 1.9 & - / 2.3 & 4.4 / 1.9 & - / - & 3.6 / 1.5 \\ 
\textbf{ sl } & 4.0 / 2.2 & 11.1 / 2.0 & 19.5 / 7.3 & 8.6 / 3.4 & - / 0.2 & 0.8 / 0.3 & 13.2 / 4.5 & 4.8 / 1.1 & 0.4 / 0.1 & 6.0 / 1.2 & 0.1 / 0.0 & 4.5 / 1.0 & 6.7 / 1.2 & - / 1.5 & 1.1 / 0.1 & 1.7 / 0.3 & - / - \\ \thickhline
\end{tabular}}
\caption{BLEU scores of bilingual S2S models on three test sets. The first score is either on EPST or \voxpop{} data, and EPST score is underscored. The second score is on FLEURS data.}
\label{tab:blg}
\end{table*}

A large-scale evaluation is launched covering $272$ mined languages directions, and bilingual models are trained for each direction to establish baseline results in speech-to-speech translation.

\autoref{tab:blg} summarizes performance of bilingual S2ST models on three test sets. In each direction, the first BLEU score is for European Parliament domain, either EPST or \voxpop{} set. EPST BLEU is underlined to be distinguished from \voxpop{} BLEU.  The second score is for Wikipedia domain, i.e., FLEURS test data.

\textbf{Bilingual results}. Empirically we find that translations into high-resource languages such as en, es and fr outperform those into low-resource languages such as lt and sl based on the language resource sizes in \autoref{stats}. Another observation is the performance difference across test domains, i.e., BLEU on FLEURS is lower than that on EPST and \voxpop{} data likely because of the domain mismatch between train ad test data. 

It is also found that translation results are not symmetric for some language pairs, for example, ro-en has a BLEU of $22.6$ while en-ro BLEU is only $7.6$ on EPST. Besides different complexity levels of target languages and test sets, such asymmetry also results from the dependency of BLEU score on the  speech synthesis quality of the vocoder and transcription quality of the ASR model. For languages whose vocoder and ASR models are not well trained, they are likely to receive low BLEU scores.
In this case, Romanian vocoder and ASR are not as strong as English models as reflected by its higher word error rate in speech resynthesis as reported in Appendix \ref{app:vocoder}.

\section{Multilingual Speech-to-Speech Translation}
\label{sec:mlg}

Multilingual modeling has been explored for the tasks of language models and machine translation, demonstrating knowledge transfer among languages. However, to our best knowledge, there are very few studies of
multilingual speech-to-speech translation, partially due to the lack of multilingual speech-to-speech resources. With the massively multilingual data we have mined, we are able to explore multilingual S2ST training. 

In this work, we focus on many-to-English translation, studying the translation from $6$ Slavic languages to English in \autoref{subsec:slavic_to_en} and the translation from all $16$ languages in \SM{} to English in \autoref{subsec:all_to_en}. English-to-many or many-to-many translation are left to future work.

\begin{table*}[htbp!]
\centering
\resizebox{0.9\textwidth}{!}{
\begin{tabular}{ccccc|cccccc}
\thickhline
& \multicolumn{4}{c|}{Bilingual} & \multicolumn{6}{c}{Multilingual} \\ \hline
 & \multicolumn{2}{c}{70M} & \multicolumn{2}{c|}{260M} & \multicolumn{2}{c}{70M} & \multicolumn{2}{c}{260M} & \multicolumn{2}{c}{424M}  \\
 & EP / VP & FLEURS & EP / VP & FLEURS & EP / VP & FLEURS & EP / VP & FLEURS & EP / VP & FLEURS \\ \hline
\textbf{cs} & 22.7 & 4.2 & 24.7 & 11.2 & 19.7 & 2.3 & 27.5 & 13.7 & 25.3 & 10.2 \\ 
\textbf{hr} & - & 7.7 & - & 4.6 & - & 3.1  & - & 12.8 & - & 9.2 \\ 
\textbf{lt} & 3.1 & 0.9 & 0.2 & 0.0  & 2.8 & 0.3 & 14.7 & 4.8 & 10.7 & 3.3 \\ 
\textbf{pl} & \underline{4.9} & 4.9 & \underline{17.6} & 7.7  & \underline{14.4} & 1.9 & \underline{19.9} & 9.5 & \underline{16.4} & 6.9 \\ 
\textbf{sk} & 21.4 & 5.5 & 24.4 & 11.0 & 18.9 & 4.1 & 27.2 & 15.4 & 24.9 & 11.1 \\ 
\textbf{sl} & 19.5 & 7.3 & 16.9 & 4.7 & 14.6 & 3.1 & 22.9 & 10.7 & 21.0 & 7.6 \\ \hline
\textbf{avg} & 14.3 & 5.1 & 16.8 & 6.5 & 14.1 & 2.5 & 22.4 & 11.2 & 19.7 & 8.1 \\
\thickhline 
\end{tabular}}
\caption{BLEU of Slavic-to-English multilingual \tless{} across domains (for EP / VP column, underlined scores are on EPST data, and others on \voxpop{} data).}
\label{tab:mlg_slavic}
\end{table*}

\begin{table*}[htbp!]
\centering
\resizebox{0.9\textwidth}{!}{
\begin{tabular}{ccc|cc|cc}
\thickhline
 & \multicolumn{2}{c|}{Bilingual (1.2B)} & \multicolumn{2}{c|}{Multilingual Dense(1.2B)} &  \multicolumn{2}{c}{Multilingual MoE-GShard64 (4.3B)} \\ \hline
 & EP / VP & FLEURS & EP / VP & FLEURS & EP / VP & FLEURS \\ \hline
\textbf{cs} & 28.3 & 17.8 & 29.7 & 18.2 & 30.6 & 19.3 \\
\textbf{hr} & - & 12.1 & - & 17.1 & - & 17.6 \\
\textbf{lt} & 0.0 & 0.0 & 20.9 & 9.6 & 22.2 & 10.2 \\
\textbf{pl} & \underline{17.4} & 7.4 & \underline{21.1} & 12.9 & \underline{21.4} & 12.6 \\
\textbf{sk} & 24.7 & 14.5 & 30.8 & 19.3 & 31.8 & 20.0 \\
\textbf{sl} & 20.1 & 8.5 & 27.4 & 14.0 & 29.1 & 13.0 \\ \hline
\textbf{avg} & 18.1 & 10.1 & 26.0 & 15.2 & 27.0 & 15.5 \\ \thickhline
\end{tabular}}
\caption{BLEU of Slavic-to-English multilingual \xm{} models in different domains (for EP / VP column, underlined scores are on EPST data, and others on \voxpop{} data).}
\label{tab:mlg_xm_slavic}
\end{table*}

Multilingual models in our experiments include
\begin{itemize}
    \item \textbf{\tless{}}. The same model that we use for bilingual evaluation is reused in the multilingual experiments. The bilingual \tless{} has $70$M parameters. Given the increased amount and diversity of multilingual data, we increase the model size for larger model capacity, trying multilingual models with $70$M, $260$M and $424$M parameters respectively.
    \item \textbf{\xm{}}. Inspired by the recent finding that crossmodal pre-training is beneficial for speech translation \cite{DBLP:journals/corr/abs-2204-02967}, we introduce \xm{} to multilingual training, whose encoder is initialized from pre-trained XLS-R model with 1B parameters \cite{DBLP:journals/corr/abs-2111-09296} and decoder is initialized from a unit decoder pre-trained in an mBART style \cite{DBLP:journals/corr/abs-2204-02967}. With multilingual speech-to-unit data, the model is further finetuned to minimize the cross-entropy loss in unit prediction.
    \item \textbf{\xm{} with Sparsity}. Sparse modeling, in particular Mixture-of-Experts (MoE), has been widely studied in multilingual machine translation as an efficient approach to encourage knowledge transfer and mitigate interference among languages. MoE increases the number of parameters of the model in magnitude without sacrificing computation efficiency. In this work, we experiment with two variants of sparse modeling, GShard \cite{lepikhin2020gshard} and Base Layer \cite{lewis2021base}.

\textbf{GShard.} GShard is a sparse scaling technique for transformer proposed in \cite{lepikhin2020gshard}. We replace every other transformer layer with an MoE layer. FFN layers in an MoE transformer layer are shared across experts. One GPU could host one or more experts and a learned gating function is used to determine which expert a token is routed to. Similar approaches have also been adopted recently in the multilingual text translation area to address language interference issues \cite{nllb}.  We apply GShard architecture on the decoder of \xm{}, to expand the capacity without the need to retrain larger unit mBART models. All expert weights are initialized with unit mBART. 

\textbf{Base Layer.} Base Layer is a variant of the sparse model proposed by \cite{lewis2021base}, which applies a balanced assignment to expert layers. It formulates token-to-expert
allocation as a linear assignment problem, encouraging each expert to receive an equal number of tokens. The assignment scheme improves efficiency by balancing compute loads, and simplifies training without any new hyperparameters or auxiliary losses. In our experiments, we add one additional Base Layer in the middle (the 7th layer) of decoder. The weights of Base Layer experts are randomly initialized. 

\end{itemize}


Appendix \ref{app:mlg} provides more details about these multilingual model configurations.

\subsection{Slavic-to-English Translation}
\label{subsec:slavic_to_en}


The six Slavic languages include Czech (cs), Croatian (hr), Lituanian (lt), Polish (pl), Slovak (sk), and Slovenian (sl). In the multilingual setting, mined data into English at the threshold of $1.06$ are combined from each Slavic language as the train set. As for the evaluation of multilingual models, we report ASR BLEU in each direction, respectively.

We first extend \tless{} from the bilingual to multilingual setting. 
For \tless{}s with different parameter sizes, their multilingual as well as bilingual translation results are presented in \autoref{tab:mlg_slavic}. 

\textbf{Results}. With the \tless{} size fixed as $70$M, multilingual training hurts the performance of most languages compared with bilingual training. This is due to the insufficient model capacity, and the language interference is reflected by an average of $-2.6$ BLEU in FLEURS. We increase model parameters to $260$M in both bilingual and multilingual settings. With a larger model capacity, bilingual models achieve gains in high-resource languages including cs, pl and sk, while suffering from performance loss in low-resource directions such as hr, lt and sl.

\begin{table*}[htbp!]
\centering
\resizebox{0.6\textwidth}{!}{
\begin{tabular}{ccc|cc|cc}
\thickhline
 & \multicolumn{2}{c|}{Dense (1.2B)} & \multicolumn{2}{c|}{MoE-GShard64 (4.3B)} & \multicolumn{2}{c}{Base Layer (1.7B)}  \\ \hline
 & EP / VP & FLEURS & EP / VP & FLEURS & EP / VP & FLEURS  \\ \hline
\textbf{cs} & 29.9 & 18.7 & 30.9 & 18.2 & 29.9 & 17.3  \\
\textbf{de} & \underline{18.8} & 19.0 & \underline{19.3} & 20.3 & \underline{19.4} & 19.5  \\
\textbf{es} & \underline{22.8} & 15.2 & \underline{23.3} & 15.9 & \underline{22.9} & 14.9 \\
\textbf{et} & - & 16.7 & - & 16.7 & - & 16.4 \\
\textbf{fi} & 26.8 & 14.1 & 28.2 & 14.0 & 28.5 & 13.9 \\
\textbf{fr} & \underline{23.5} & 18.3 & \underline{24.1} & 18.9 & \underline{23.4} & 18.2 \\
\textbf{hr} & - & 16.6 & - & 16.8 & - & 16.3 \\
\textbf{hu} & 20.2 & 12.0 & 21.3 & 12.5 & 20.5 & 12.1 \\
\textbf{it} & 36.3 & 16.2 & 37.8 & 14.9 & 37.4 & 14.0 \\
\textbf{lt} & 21.9 & 9.8 & 23.8 & 10.3 & 23.4 & 10.0 \\
\textbf{nl} & \underline{21.4} & 16.4 & \underline{22.1} & 17.3 & \underline{21.5} & 16.6 \\
\textbf{pl} & \underline{21.2} & 12.4 & \underline{21.3} & 13.4 & \underline{20.9} & 12.5 \\
\textbf{pt} & \underline{23.8} & 21.8 & \underline{24.2} & 22.3 & \underline{23.8} & 21.1 \\
\textbf{ro} & \underline{25.1} & 19.7 & \underline{25.0} & 19.8 & \underline{25.3} & 19.0 \\
\textbf{sk} & 30.8 & 19.6 & 32.2 & 18.2 & 31.5 & 18.4 \\
\textbf{sl} & 28.3 & 13.7 & 29.9 & 13.7 & 28.8 & 13.5  \\ \hline
\textbf{avg} & 25.1 & 16.3 & 26.0 & 16.5 & 25.5 & 15.9 \\ \thickhline
\end{tabular}}
\caption{BLEU of All-to-English multilingual models in different domains (for EP / VP column, underlined scores are on EPST data, and others on \voxpop{} data).} 
\label{tab:mlg_all}
\end{table*}

Given model sizes of $260$M, we observe consistent gains of multilingual models over the bilingual models across different language directions and test domains. An average gain of $5.6$ BLEU is achieved in EP/VP and the gain of $4.7$ BLEU in FLEURS. It demonstrates the positive transfer enabled by multilingual training. As the multilingual model size continues to increase to $424$M, we don't observe further gains likely due to the bottleneck of training data amount.

Pre-training has been shown to be beneficial for speech-to-speech translation \cite{DBLP:journals/corr/abs-2204-02967}. \xm{} leveraging pre-trained modules is also trained on Slavic-to-English data, and results are reported in \autoref{tab:mlg_xm_slavic}.
Comparing against bilingual \tless{} (70M), bilingual \xm{} outperforms it in all directions except lt-en. The gain in EP/VP is $3.8$ BLEU on average, and a larger gain of $5.0$ BLEU is achieved in FLEURS. Multilingual training brings further gains to \xm{} with +$7.9$ and +$5.1$ BLEU over bilingual training in EP/VP and FLEURS test set respectively. 

Comparing against dense \xm{}, GShard with 64 experts has +$1.0$ BLEU gains on average over 5 directions on EP/VP, and +$0.3$ BLEU gains for FLEURS. We believe it's due to a phenomena mentioned in \cite{zoph2022designing} that MoE specializes in multilingual settings but not by language. GShard in our setting brings larger improvements to in-domain test set.

Overall the best Slavic-to-English translation is achieved by \xm{} with GShard trained in multilingual setting. This demonstrates that pre-training, model sparsity and multilinguality are of help to speech-to-speech translation.



\subsection{All-to-English Translation}
\label{subsec:all_to_en}

We move forward to a larger-scale multilinguality by extending from Slavic language family to all languages in \SM{}. We adopt the best models in Slavic-to-English translation, i.e., multilingual \xm{} with both dense and sparse architectures. In terms of sparse modeling, we try both GShard and Base Layer in all-to-English translation.

\textbf{Results}. Compared with \xm{} (1.2B) dense model, MoE-GShard64 (4.3B) with the same forward computation time brings gains of +$0.9$ and +$0.2$ BLEU to EP/VP and FLEURS respectively. Similar to our findings in Slavic-to-English setting, increasing the capacity with sparse modeling benefits more on in-domain (EP/VP) than out-of-domain FLEURS test set.

With the sparse architecture of XM Transformer with GShard, all-to-English model shows +$0.6$ and -$0.4$ BLEU difference compared with Slavic-to-English model on EP/VP and FLEURS respectively, averaged over Slavic languages. Multilingual sparse model benefits from the additional in-domain data in other languages when evaluated in EP/VP domain, while sees performance degradation in out-of-domain data.  

The other sparse variant, Base Layer (1.7B) performs comparably to the dense \xm{}, with an average of +$0.4$ BLEU in EP/VP test sets and -$0.4$ BLEU in FLEURS. The sparsity in Base Layer does not bring obvious gains to all-to-English translation. This is likely because we only add one Base Layer to the decoder with a small expert size. The number of increased model parameters is only $0.5$B in Base Layer, while it is $3.1$B in GShard. As suggested by \cite{lewis2021base}, the Base Layer performance might improve with more GPUs and a larger expert size.


\section{Limitations}

The HuBERT model is critical to the speech-to-speech translation performance as its extracted units are used by both speech-to-unit model and vocoder. We have not explored the optimal strategy of multilingual HuBERT training. One research question is how to choose a group of languages that a multilingual HuBERT model is trained on. For example, it is arguable whether Lithuanian (lt) should be included in Slavic or Uralic family. Other questions could be whether a larger HuBERT with more model capacity should be used and how we should deal with language imbalance in multilingual training.

We provide benchmark results of bilingual speech translation with mined data selected by heuristics. One of our future directions is to come up with a better strategy of mined data selection to improve translation performance and training efficiency. 

As mentioned in our results analysis, the reported BLEU scores are heavily dependent on the ASR quality, which may not reflect the speech translation performance accurately. Future directions could be improving ASR quality or exploring other evaluation metrics without reliance on ASR models.


\section{Conclusion}

In this paper, we introduce a large-scale multilingual speech-to-speech corpus mined from \voxpop{}. It is the largest resource of speech alignments with a coverage of $\nbLangs{}$ languages. We perform an extensive evaluation of the mined parallel speech, showing good quality of the speech alignments. Multilingual speech-to-speech models can be efficiently trained on this corpus and we suggest different methods, like sparse scaling using Mixture-of-Experts, to further boost translation performance in the multilingual setting.

\bibliography{anthology,custom}
\bibliographystyle{acl_natbib}

\clearpage
\appendix

\section{Appendix}
\label{sec:appendix}

\subsection{Similarity search on CoVoST}
We compared our similarity search results with previous work \cite{duquenne2021multimodal} in \autoref{tab:simsearch_previous}. We notice that our new speech encoders have lower error rates compared to previous work.

\begin{table}[h!]
\centering
\small
\begin{tabular}{llll}
\thickhline
\textbf{Audio vs. en translations} & \textbf{de} & \textbf{es} & \textbf{fr}  \\
Previous work                   & 3.36        & 1.66        & 2.05         \\
This work                          & \textbf{3.27}        & \textbf{1.26}        & \textbf{1.55}   \\
\thickhline
\end{tabular}
 \caption{Similarity search error rates (in \%) on \covost{} test set.}
 \label{tab:simsearch_previous}
\end{table}


\subsection{HuBERT}
\label{app:hubert}

We train a multilingual HuBERT model for each family on the collection of speech in each component language as shown in \autoref{tab:fam}. The HuBERT model consists of $7$ convolutional layers and $12$ Transformer encoder layers. Each encoder layer has 12 attention heads, the embedding dimension of $768$ and the forward dimension of $3072$. Models are trained for $3$ iterations, and in each iteration pseudo-labels are prepared as the training target for utterances. In the first iteration, the target labels are MFCC features. In the second iteration, we extract speech features from the $6$-th layer of the trained HuBERT model and apply $k$-means clustering to derive a set of $500$ labels. In the third iteration, speech features from the $9$-th layer are clustered into $500$ labels. Lastly after these three iterations, we try feature extraction from different layers including layer $10$, $11$ and $12$ of trained HuBERT. As for feature clustering, we also try different numbers of clusters, $800$, $1000$ and $1200$, to derive multiple sets of target units.

To choose the optimal setup, we launch a resynthesis evaluation to select the HuBERT layer to extract speech features and the number of $k$-means clusters. We train a vocoder on each set of target units, i.e., vocoder takes the units and synthesizes target speech. The synthesized speech is sent to off-the-shelf ASR models, and Word Error Rate (WER) is reported to measure the speech quality. The set of target units is selected if the corresponding vocoder achieves lowest WER.

\subsection{ASR models}
\label{app:asr}

We use ASR models publicly released on HuggingFace to transcribe the generated speech in order to calculate WER or BLEU scores in comparison with ground truth texts. ASR models used in our evaluation are listed in \autoref{tab:asr}.

\begin{table*}[htbp!]
\resizebox{1.0\textwidth}{!}{
\begin{tabular}{c|cc}
\thickhline
\textbf{Lang} & \textbf{cs} & \textbf{de} \\ 
\textbf{ASR} & comodoro/wav2vec2-xls-r-300m-cs-250 & jonatasgrosman/wav2vec2-xls-r-1b-german \\ \hline
\textbf{Lang} & \textbf{et} & \textbf{fi} \\
\textbf{ASR} & RASMUS/wav2vec2-xlsr-1b-et & jonatasgrosman/wav2vec2-large-xlsr-53-finnish \\ \hline
\textbf{Lang} & \textbf{hr} & \textbf{hu} \\
\textbf{ASR} & classla/wav2vec2-xls-r-parlaspeech-hr & jonatasgrosman/wav2vec2-large-xlsr-53-hungarian \\ \hline
\textbf{Lang} & \textbf{it} & \textbf{lt} \\
\textbf{ASR} & jonatasgrosman/wav2vec2-large-xlsr-53-italian & sammy786/wav2vec2-xlsr-lithuanian \\ \hline
\textbf{Lang} & \textbf{nl} & \textbf{pl} \\
\textbf{ASR} & jonatasgrosman/wav2vec2-xls-r-1b-dutch & jonatasgrosman/wav2vec2-xls-r-1b-polish \\ \hline
\textbf{Lang} & \textbf{pt} & \textbf{ro} \\
\textbf{ASR} & jonatasgrosman/wav2vec2-xls-r-1b-portuguese & gigant/romanian-wav2vec2 \\ \hline
\textbf{Lang} & \textbf{sk} & \textbf{sl} \\
\textbf{ASR} & anuragshas/wav2vec2-xls-r-300m-sk-cv8-with-lm & anuragshas/wav2vec2-xls-r-300m-sl-cv8-with-lm \\ \thickhline
\end{tabular}}
\caption{HuggingFace ASR models for each language.}
\label{tab:asr}
\end{table*}

\subsection{Vocoder}
\label{app:vocoder}

Vocoders are trained to synthesize speech from a given sequence of units. The train sets are speech data from CSS10, \voxpop{} and Common Voice. As mentioned before, units are derived from HuBERT models for these speech. \autoref{tab:wer} summarizes WER of ASR models, which reflects the transcription quality in each language. Besides, we report the training dataset, vocoder WER of synthesized speech from vocoders, and here we include the vocoder results obtained from the optimal HuBERT layer and $k$-means cluster size. Layer $11$ is the best HuBERT layer for feature extraction in all languages, and most languages have the best $k$-means size of $1000$ except Italian (it) whose best label size is $800$.

\begin{table*}[htbp!]
\resizebox{1.0\textwidth}{!}{
\begin{tabular}{ccccc||ccccc}
\thickhline
Lang & Data & ASR WER & HuBERT & Vocoder WER & Lang & Data & ASR WER & HuBERT & Vocoder WER \\ \hline
de & CSS10 & 0.10 & \begin{tabular}[c]{@{}c@{}}Germanic HuBERT\\ layer 11, km 1000\end{tabular} & 0.16 & nl & CSS10 & 0.19 & \begin{tabular}[c]{@{}c@{}}Germanic HuBERT\\ layer 11, km 1000\end{tabular} & 0.27 \\ \hline
fi & CSS10 & 0.02 & \begin{tabular}[c]{@{}c@{}}Uralic HuBERT\\ layer 11, km 1000\end{tabular} & 0.15 & hu & CSS10 & 0.21 & \begin{tabular}[c]{@{}c@{}}Uralic HuBERT\\ layer 11, km 1000\end{tabular} & 0.21 \\ \hline
et & \begin{tabular}[c]{@{}c@{}}Common\\ Voice\end{tabular} & 0.14 & \begin{tabular}[c]{@{}c@{}}Uralic HuBERT\\ layer 11, km 1000\end{tabular} & 0.44 & it & \voxpop{} & 0.23 & \begin{tabular}[c]{@{}c@{}}Uralic HuBERT\\ layer 11, km 800\end{tabular} & 0.27 \\ \hline
pt & \begin{tabular}[c]{@{}c@{}}Common\\ Voice\end{tabular} & 0.06 & \begin{tabular}[c]{@{}c@{}}Uralic HuBERT\\ layer 11, km 1000\end{tabular} & 0.31 & ro & \voxpop{} & 0.42 & \begin{tabular}[c]{@{}c@{}}Uralic HuBERT\\ layer 11, km 1000\end{tabular} & 0.50 \\ \hline
cs & \voxpop{} & 0.15 & \begin{tabular}[c]{@{}c@{}}Slavic HuBERT\\ layer 11, km 1000\end{tabular} & 0.23 & pl & \voxpop{} & 0.14 & \begin{tabular}[c]{@{}c@{}}Slavic HuBERT\\ layer 11, km 1000\end{tabular} & 0.23 \\ \hline
hr & \voxpop{} & 0.21 & \begin{tabular}[c]{@{}c@{}}Slavic HuBERT\\ layer 11, km 1000\end{tabular} & 0.29 & lt & \voxpop{} & 0.38 & \begin{tabular}[c]{@{}c@{}}Slavic HuBERT\\ layer 11, km 1000\end{tabular} & 0.57 \\ \hline
sk & \voxpop{} & 0.28 & \begin{tabular}[c]{@{}c@{}}Slavic HuBERT\\ layer 11, km 1000\end{tabular} & 0.41 & sl & \voxpop{} & 0.37 & \begin{tabular}[c]{@{}c@{}}Slavic HuBERT\\ layer 11, km 1000\end{tabular} & 0.46 \\ \thickhline
\end{tabular}}
\caption{Benchmark results of ASR models and vocoder resynthesis.}
\label{tab:wer}
\end{table*}

As shown in \autoref{tab:wer}, ASR models are of good quality for high-resource languages such as de, fi and pt, while suffering from high error rates in languages such as ro, lt and sl. It is expected to have higher vocoder WER than ASR WER since the former is for synthesized speech. By measuring the gap between the two error rates, we can tell how good a vocoder is and also infer the quality of HuBERT units. For et, pt and lt, the gaps are obviously larger than other languages. It not surprising since we do not have much good-quality speech data for these languages, for example, there is only around $10$-hour noisy speech from Common Voice for et and pt vocoder training.

\section{Multilingual Speech-to-Speech Translation}
\label{app:mlg}

We provide details of models and experiment setups in multilingual speech-to-speech translation.

\subsection{Slavic-to-English Translation}
\textbf{\tless{}}. \tless{} (260M) has a speech encoder with $4$ convolution layers and $12$ Transformer encoder layers with the embedding dimension of $1024$ and the forward dimension of $4096$. It has two unit decoders with $6$ and $2$ Transformer decoder layers for target and source unit prediction respectively. The target unit decoder has the embedding dimension of $1024$ and the forward dimension of $4096$, and the source unit decoder's dimensions are $256$ and $2048$.

For the \tless{} (424M), its speech encoder contains $6$ convolution layers and $16$ Transformer encoder layers with the embedding dimension of $1024$ and the forward dimension of $4096$. It has two unit decoders with $12$ and $2$ Transformer decoder layers for target and source unit prediction respectively. The target unit decoder has the embedding dimension of $1024$ and the forward dimension of $4096$, and the source unit decoder's dimensions are $256$ and $2048$.

\textbf{\xm{}}. \xm{} (1.2B) is initialized from XLS-R encoder with $7$ convolution layers and $48$ Transformer encoder layers with the embedding dimension of $1280$ and the forward dimension of $5120$. Its unit decoder is initialized from a pre-trained mbart-style decoder with $12$ layers, embedding dimension of $1024$ and forward dimension of $4096$.

\subsection{All-to-English Translation}

\textbf{\xm{}-GShard}. XM Transfomer (1.2B) is initialized with the same XLS-R encoder and unit decoder used in Slavic-English. On the decoder side of \xm{}-GShard, each expert is initialized with the same unit decoder. We set MoE frequency as 2, i.e., every other transformer layer is an MoE layer. 

\textbf{\xm{}-Base Layer}. For our \xm{} with Base Layer sparsity (1.7B), the encoder is initialized with the same XLS-R encoder, and the dense layers of the decoder is initialized with the same unit decoder as GShard. We add an additional Base Layer which is randomly initialized as the $7$th layer of decoder. There is one expert in each GPU and we used $64$ GPUs in our experiments, which means we have $64$ Base Layer experts in total. 

\end{document}